\pdfoutput=1
\documentclass[11pt]{article}

\usepackage[]{ACL2023}

\usepackage{times}
\usepackage{latexsym}
\usepackage{graphicx}
\usepackage{multirow}

\usepackage[T1]{fontenc}
\usepackage[utf8]{inputenc}

\usepackage{microtype}

\usepackage{inconsolata}

\title{ChartSumm: A Comprehensive Benchmark for Automatic Chart Summarization of Long and Short Summaries}

\author{
  Raian Rahman\textsuperscript{1},
  Rizvi Hasan\textsuperscript{1},
  Abdullah Al Farhad\textsuperscript{1},
  Md Tahmid Rahman Laskar\textsuperscript{2,3}, \\
  \bf{Md. Hamjajul Ashmafee\textsuperscript{1}},
  \bf{Abu Raihan Mostofa Kamal\textsuperscript{1}}\\
  \textsuperscript{1}Department of Computer Science and Engineering, Islamic University of Technology, Bangladesh\\
  \textsuperscript{2}School of Information Technology, 
   York University, Canada\\
  \textsuperscript{3}Dialpad Canada Inc., Canada \\
  \{raianrahman, rizvihasan, alfarhad, ashmafee, raihan.kamal\}@iut-dhaka.edu, 
  tahmid20@yorku.ca
}

\begin{document}
\maketitle
\begin{abstract}
Automatic chart to text summarization is an effective tool for the visually impaired people along with providing precise insights of tabular data in natural language to the user. A large and well-structured dataset is always a key part for data driven models. In this paper, we propose ChartSumm: a large-scale benchmark dataset consisting of a total of 84,363 charts along with their metadata and descriptions covering a wide range of topics and chart types to generate short and long summaries. Extensive experiments with strong baseline models show that even though these models generate fluent and informative summaries by achieving decent scores in various automatic evaluation metrics, they often face issues like suffering from hallucination, missing out important data points, in addition to incorrect explanation of complex trends in the charts. We also investigated the potential of expanding ChartSumm to other languages using automated translation tools. These make our dataset a challenging benchmark for future research.
\end{abstract}

\section{Introduction}
\label{intro}
Automatic chart summarization is a task where the goal is to describe important data points and trends in a chart in natural language. Chart summaries are helpful to better interpret the chart, making it useful for the visually impaired people as well as to improve the performance of different information retrieval algorithms \citep{obeid-hoque-2020-chart,carenini2013user,li2013towards}.

Scarcity of large scale well defined datasets with chart image, metadata and well described summaries is a major challenge in automatic chart summarization. To our best knowledge, there are only four datasets \citep{obeid-hoque-2020-chart, zhu2021autochart, hsu-etal-2021-scicap-generating, kanthara2022chart} available for the chart to text summarization task, making this task a low resource problem. Among these datasets, three of them \citep{obeid-hoque-2020-chart, zhu-etal-2021-autochart, kanthara2022chart} contain chart images with metadata and well defined summaries while the SciCAP dataset \citep{hsu-etal-2021-scicap-generating} only contains chart images and captions.

In this work, we address the scarcity of public datasets in the automatic chart summarization task. We propose ``\textbf{ChartSumm}'', a large scale dataset for chart to text summarization comprising of $84,363$ chart images with corresponding chart metadata and summaries. 
(see Figure \ref{fig:example-intro} for an example). 
We also propose two test sets based on the summary length.
In this paper, our major contributions are summarized below: 

\begin{figure}[t!]
    \centering
    \includegraphics[width = \columnwidth]{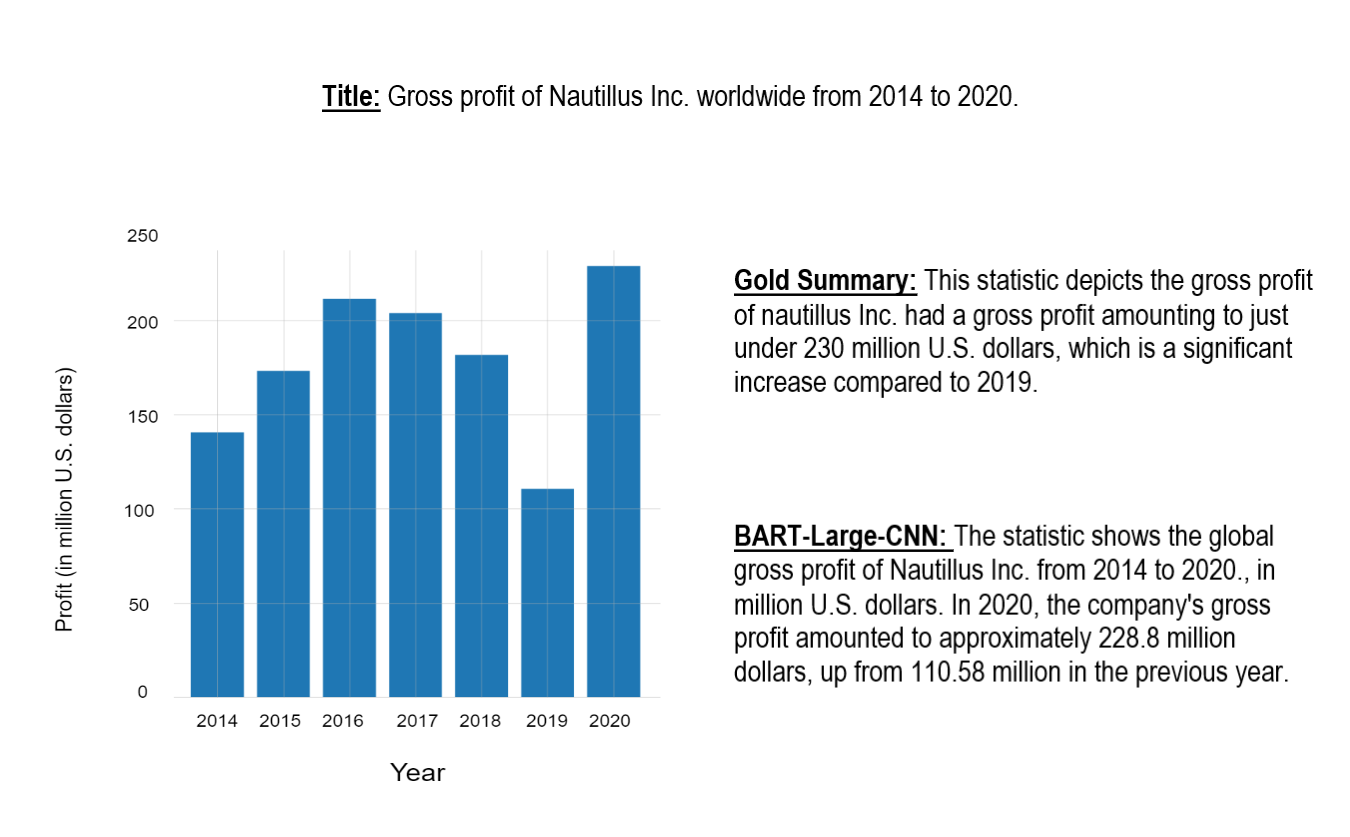}
    \caption{An example chart-summary pair from our proposed dataset.}
    \label{fig:example-intro}
\end{figure}

(i) Proposing a new benchmark dataset for the automatic chart summarization task. To our best knowledge, our ChartSumm dataset is currently the largest dataset proposed for this task. Meanwhile, we also introduce two different test sets to separately compare the performance on generating short and long summaries.

\begin{table*}[t!]
\centering
\resizebox{\textwidth}{!}{%
\begin{tabular}{llllll}
\hline
\multicolumn{1}{l}{\textbf{Dataset Name}} & \multicolumn{1}{l}{\textbf{Task}}   & \multicolumn{1}{l}{\textbf{Data Source}}                 & \multicolumn{1}{l}{\textbf{Formulation}} & \multicolumn{1}{l}{\textbf{Summary Type}} & \multicolumn{1}{l}{\textbf{Example Count}} \\ \hline
SciCap \citep{hsu-etal-2021-scicap-generating} & Image → Text caption                                                                                                             & Scientific Papers                                         & \begin{tabular}[l]{@{}l@{}}Chart Image\\ Captions\end{tabular}               & Short captions from scientific paper figures                                                                                            & 290,000                          \\ \hline
Chart2Text \citep{obeid-hoque-2020-chart}                       & Table → Text description                                                                                                         & Statista                                                  & \begin{tabular}[c]{@{}l@{}}Chart Image\\ Metadata\\ Description\end{tabular} & Descriptive human written summary                                                                                                       & 8,305                             \\ \hline
AutoChart \citep{zhu2021autochart}                        & Table → Text description                                                                                                         & Different public repository                               & \begin{tabular}[c]{@{}l@{}}Chart Image\\ Metadata\\ Description\end{tabular} & Automatic template generated summary                                                                                                    & 23,543                            \\ \hline
Chart-To-Text \citep{kanthara2022chart}       & \begin{tabular}[c]{@{}l@{}}1. Table → Text description\\ 2. Image → Extracted metadata using OCR → Text description\end{tabular} & \begin{tabular}[c]{@{}l@{}}Statista\\ Pew\end{tabular}    & \begin{tabular}[c]{@{}l@{}}Chart Image\\ Metadata\\ Description\end{tabular} & \begin{tabular}[c]{@{}l@{}}Descriptive human written summary from\\ pew and statista\end{tabular}                                       & 44,085                            \\ \hline
ChartSumm (ours)                 & Table → Text description                                                                                                         & \begin{tabular}[c]{@{}l@{}}Statista\\ Knoema\end{tabular} & \begin{tabular}[c]{@{}l@{}}Chart Image\\ Metadata\\ Description\end{tabular} & \begin{tabular}[c]{@{}l@{}}System generated short summary from Knoema\\ and descriptive human written summary from staista\end{tabular} & 84,363                            \\ \hline
\end{tabular}%
}
\caption{Comparison between existing datasets and our proposed dataset.}
\label{comparisonDataset}
\end{table*}

(ii) Conducting a series of experiments using strong baselines to  demonstrate how models trained on our dataset have better generalization capability than other existing datasets. In addition, we identify the limitations of the state-of-the-art models in our proposed dataset. Furthermore, we also explore the scope of expanding our dataset to other languages through translation and evaluate the performance in a human-annotated test set in the Bengali language. To our best knowledge, this is the first work that investigated the Chart Summarization task in any languages except English. The dataset and codes are available at \href{https://github.com/pranonrahman/ChartSumm}{https://github.com/pranonrahman/ChartSumm}.
% As a secondary contribution, we will make our dataset publicly available. 

\section{Related Work}

Existing Chart-To-Text summarization systems generate summaries from either the chart image \cite{hsu-etal-2021-scicap-generating} or the chart metadata  \cite{gong2019enhanced,obeid-hoque-2020-chart,kanthara2022chart}. Before the advent of deep learning, most early work utilized a two stage approach that applied content selection using different statistical tools in the first step followed by generating summaries using pre-defined templates \citep{reiter-2007-architecture,zhu-etal-2021-autochart}. However, predefined template-based architectures frequently lack generality and fail to capture complex trends in data. In recent years, deep learning-based techniques have gained significant attention \citep {gong2019enhanced,obeid-hoque-2020-chart,inproceedings,zhu2021autochart,hsu-etal-2020-efficient,zhou2021reverse, dadhich2021barchartanalyzer, sreevalsan2021tensor,luo2021chartocr,8989823,kanthara2022chart} due to their superior performance over the template-based approaches. Nonetheless, due to the lack of Chart-To-Text summarization datasets, not only that the models proposed for this task require improvements, the generalized effectiveness of these models is also yet to be investigated. 

Among the four benchmark Chart-To-Text summarization datasets that are publicly available, the Chart2Text \citep{obeid-hoque-2020-chart} summarization dataset is the first dataset proposed for this task that includes 8,305 samples collected from the Statista\footnote{\href{https://www.statista.com/}{https://www.statista.com/}}. %: a statistical analysis website containing chart images, chart metadata, and human-written descriptions of the chart.
However, the size of this dataset is quite small and so effective data-driven methods cannot be trained on top of that.
Later, the SciCAP \citep{hsu-etal-2021-scicap-generating} data was proposed for the chart captioning task from chart images. Thus, it is not suitable for methods that can only generate summary from metadata. 
The recently proposed AutoChart \cite{zhu-etal-2021-autochart} dataset is based on some predetermined templates and so this dataset does not contain much variance in the chart descriptions. More recently, Kanthara et al., \cite{kanthara2022chart} proposed the Chart-To-Text dataset that consists of chart images, with their corresponding metadata and human written descriptions. Though this dataset is currently the largest dataset available for this task containing 44,085 charts that were collected from the Statista website and the Pew website, our proposed ChartSumm dataset is almost double in size than the Chart-To-Text dataset.

\section{The ChartSumm Dataset}
This section describes how we compile a large-scale dataset consisting of $84,363$ examples from the Knoema\footnote{\href{https://knoema.com/atlas}{https://knoema.com/atlas}} and the Statista website for the chart to text summarization task along with their analysis. % At first, we describe how we construct the dataset, followed by data analysis.

\subsection{Dataset Construction}
\textbf{Knoema:}
It is a statistical service-based online platform that contains the economic indicator of more than 200 countries. Knoema provides a short description for each statistic generated by its digital data assistant named Yodatai\footnote{\href{https://yodatai.knoema.com/}{https://yodatai.knoema.com/}} that summarizes basic information about datasets. To construct our dataset, we first crawl over 1,10,000 statistics from Knoema. Then, we filter out the statistics where the source of data is not publicly available, resulting in 43,179 publicly available statistics. Afterward, we collect the chart metadata and their corresponding short descriptive captions. Since the statistics in Knoema are shown with respect to the year, we classify each chart as a simple line chart. The title and caption of the chart are then tokenized while we remove white spaces and newlines using stemming. We also normalize the numerical entities.

\textbf{Statista:}
It is also an online platform where statistics on a wide range of topics are published along with a short human-written description of the statistics. Topics in Statista include economics, marketing, industry, and opinion research. For dataset creation, at first, we crawl over $750,000$ available pages in Statista research to collect a list of $41,184$ publicly available charts along with summaries and chart metadata. Then we classify the data into simple and complex charts depending on the number of columns in the chart. Similar to the Knoema dataset, we also apply tokenization and stemming. Since many examples in the Statista dataset did not contain the x\_label, we apply the following heuristic rules to automatically classify the x\_labels as Year, Month, Day, Quarter, Country, City, and Continent: 
\begin{itemize}
    \item \textbf{Year: }If all x values were integers less than 2050 and greater than 1800 we set the x label to ``year''. 
    \item \textbf{Month: }If all x values were names of months, we set the x label to ``month''. 
     \item \textbf{Day: }If all x values were names of days(saturday, sunday ...), we set the x label to ``month''. 
    \item \textbf{Quarter: }If most x values were Q1/Q2/Q3/Q4 and optionally followed by an integer(year) we set x label as ``quarter''.
    \item \textbf{Country: }If more than 30\% x values contained values from the list of all countries collected from Wikipedia, we labeled the x label as ``country''
    \item \textbf{City: }If more than 30\% x values contained values from the list of cities collected from World City Database, we labeled the x label as ``city''
    \item \textbf{Area: }If the x labels contained the names of general areas like continent names, sub continent names, etc, we set the x label as ``area''.
    \item \textbf{NER: }We also used named entity recognition to identify some other named types such as companies, social medias etc. 
\end{itemize}

For charts where the x\_labels could not be automatically determined, the value for the x\_label is set to ``x\_label''.

To classify the charts into different types (bar/line/pie), we use ChartReader\footnote{\href{https://github.com/Cvrane/ChartReader}{https://github.com/Cvrane/ChartReader}}. The charts are then divided into simple and complex categories. We need to find \textbf{missing x labels}
since some of the scrapped data have missing x labels. So we manually identify them using the following methods.

\begin{figure}[t!]
    \centering
    \includegraphics[width=\columnwidth]{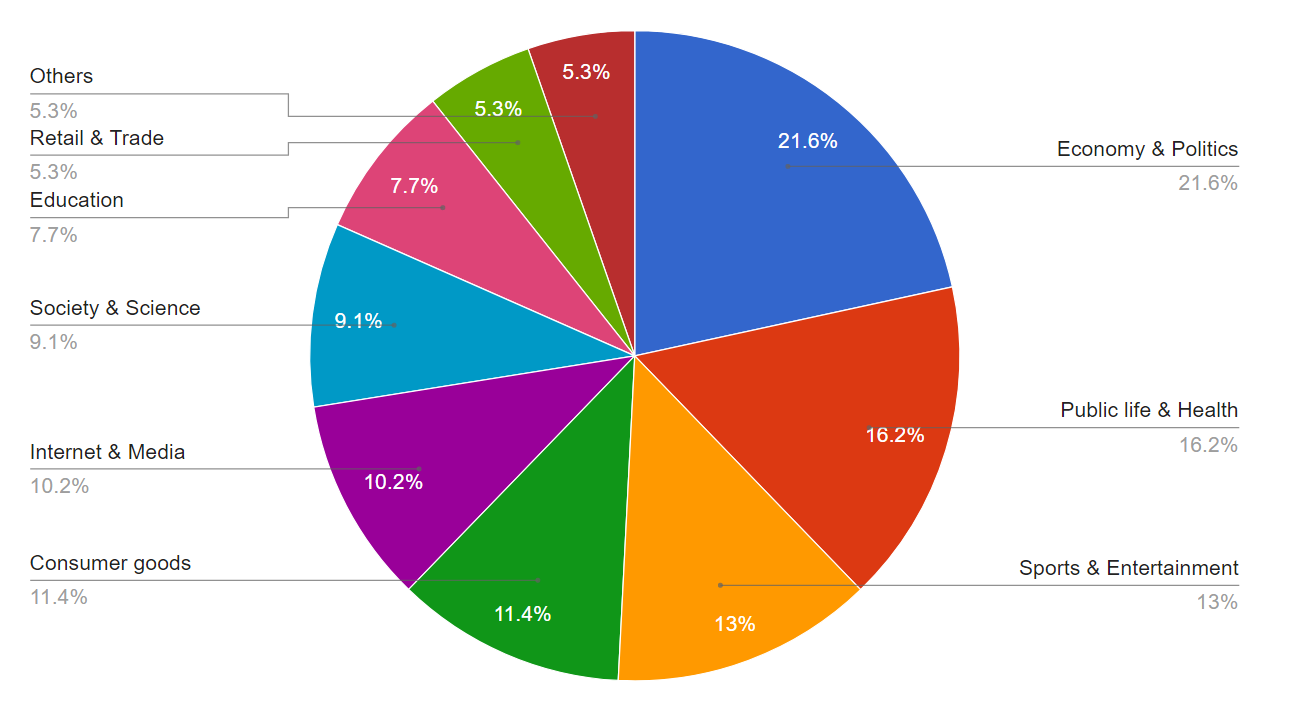}
    \caption{Topic distribution of ChartSumm}
    \label{topicDistribution}
\end{figure}

\subsection{Dataset Analysis}

In this section, we analyze our proposed ChartSumm dataset. At first,  we compare our ChartSumm dataset with some existing datasets in Table \ref{comparisonDataset}. We find the both Chart2Text \citep{obeid-hoque-2020-chart} and Chart-To-Text \citep{kanthara2022chart} collected their data from a source called Statista in which the summaries are bit descriptive and longer in length. Whereas our dataset contains both long and short summaries.

Note that our proposed dataset contains line charts, bar charts, and pie charts. For Statista. bar chart is the most common type ($64.70\%$ in statista) followed by line chart ($33.76\%$ in statista) followed by pie chart ($1.54\%$ in statista). For knoema, all charts are line charts. In Figure \ref{topicDistribution}, we show the topic distribution of our dataset. For topic modeling, we perform Latent Dirichlet allocation (LDA) \citep{10.5555/944919.944937} by creating a topic per document and a word per topic model, both of which are based on Dirichlet distributions. We find that our proposed ChartSumm dataset covers a large spectrum of topics including Economy \& politics ($21.60\%$), Society \& Science ($13.03\%$), Internet \& Media ($11.43\%$), Public life \& Health ($10.42\%$), Sports \& Entertainment ($9.14\%$), Consumer Goods ($7.71\%$), Retail \& Trade ($5.35\%$), Education ($5.32\%$), etc.

\begin{table}[t!]
\centering
\resizebox{\columnwidth}{!}{%
\begin{tabular}{|l|c|cc|cc|}
\hline
\multirow{2}{*}{\textbf{Source}} & \multirow{2}{*}{\textbf{Average Cell Count}} & \multicolumn{2}{c|}{\textbf{Average Summary Length}} & \multicolumn{2}{c|}{\textbf{Average Title Length}} \\ \cline{3-6} 
                   &       & \multicolumn{1}{c|}{\textbf{Token}} & \textbf{Chars}  & \multicolumn{1}{c|}{\textbf{Token}} & \textbf{Chars}  \\ \hline
Knoema             & 55.44 & \multicolumn{1}{c|}{34.76} & 207.69 & \multicolumn{1}{c|}{8.86}  & 57.18 \\ \hline
Statista - simple  & 13.31 & \multicolumn{1}{c|}{46.96} & 288.68 & \multicolumn{1}{c|}{9.58}  & 63.08 \\ \hline
Statista - complex & 37.95 & \multicolumn{1}{c|}{55.54} & 340.19 & \multicolumn{1}{c|}{10.54} & 67.08 \\ \hline
\end{tabular}%
}
\caption{Dataset analysis of ChartSumm.}
\label{dataAnalysis}
\end{table}

In Table \ref{dataAnalysis}, we show the average cell counts for the tables, as well as character and token counts for summaries and titles, respectively. We find that in terms of average number of tokens, the summaries of simple and complex Statista charts were about $35\%$ and $59.9\%$ longer than the Knoema charts, respectively. 
The data obtained from each source was classified into train, validation, and test sets, following an 80:10:10 split ratio. We show the number of samples in our training, validation, and test sets in table \ref{datasplit}. Since, one source of our dataset is Statista that is similar to the Chart-To-Text \cite{kantharaj-etal-2022-chart} dataset, we measure the overlaps between the data samples from Statista in our ChartSumm dataset and the Chart-To-Text dataset. For similarity measurement, we first tokenize the captions and then calculate the percentage of matched tokens. We assume two samples are exactly similar when the similarity is greater than $90\%$. Table \ref{tab:overlapping} shows that only $5,338$ captions in our dataset overlaps with the samples from Statista in Chart-To-Text.

\begin{table}[t]
\centering
\resizebox{\columnwidth}{!}{
\begin{tabular}{|l|l|l|l|}
\hline
        &  \textbf{ChartSumm} &  \textbf{Chart-To-Text} & \textbf{Overlaps} \\ \hline
Simple  & 33067     & 27868         & 4144       \\ \hline
Complex & 8338      & 6943          & 1194       \\ \hline
Total   & 41405     & 34811         & 5338       \\ \hline
\end{tabular}
}
\caption{Overlaps between Chartsumm and Chart-To-Text Statista samples.}
\label{tab:overlapping}
\end{table}

\begin{table}[t]
\centering
\resizebox{\columnwidth}{!}{%
\begin{tabular}{|c|c|c|c|c|c|c|}
\hline
% Split   & Size   \\ \hline
% Train   & 67,488 \\ \hline
% Valid-k & 4,338  \\ \hline
% Valid-s & 4,101  \\ \hline
% Test-k  & 4,338  \\ \hline
% Test-s  & 4,098  \\ \hline
% Total   & 84,363 \\ \hline
\textbf{Train-k} & \textbf{Train-s} & \textbf{Valid-k} &\textbf{Valid-s} & \textbf{Test-k} & \textbf{Test-s} & \textbf{Total} \\ \hline
34,503 & 32,985 & 4,338 & 4,101 & 4,338 & 4,098 & \textbf{84,363} \\ \hline
\end{tabular}%
}
\caption{Split distribution of ChartSumm.}
\label{datasplit}
\end{table}

\section{Experiments}

In this section, we present the baseline models that we utilize to benchmark the performance in our proposed dataset, followed by the fine-tuning process, the evaluation metrics, and the experimental results. 

\subsection{Baselines:} We use T5-Base \cite{raffel2019exploring} and BART \citep{lewis-etal-2020-bart}  as our baselines due to their effectiveness in Chart-To-Text tasks \cite{kantharaj-etal-2022-chart}. 
T5 is a large pre-trained language model trained on multiple sequence-to-sequence tasks. BART is a sequence-to-sequence model pre-trained for the language modeling task using the denoising autoencoder architecture. We fine-tune three variants of BART: (i) BART-Base, (ii) BART-Large-CNN, and (iii) BART-Large-XSUM. We implement all models using HuggingFace \cite{wolf-etal-2020-transformers}. Below we describe our model fine-tuning process.

\subsection{Fine-tuning Process}
\label{APPENDIX:finetunestage}
\begin{figure}[t]
    \centering
    \includegraphics[width = \columnwidth]{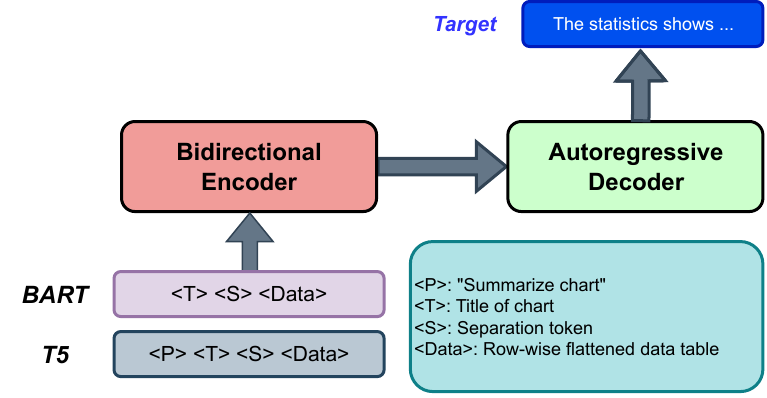}
    \caption{Fine-tuning process of baseline models}
    \label{fig:finetuneprocess}
\end{figure}

We used chart metadata (title, corresponding data table, labels) to fine-tune all four of our pre-trained baseline models. We flattened the table by rows and concatenated it with the caption of the table separated by a separator token. To mimic the pretraining process of T5, we added the prefix: ``Summarize chart: '' before each example. Figure \ref{fig:finetuneprocess} shows the fine-tuning stages.

All four baseline models were fine-tuned for $3$ epochs with a batch size of $8$. The initial learning rate during our fine-tuning was $1e-6$. We used AdamW  \cite{kingma2014adam,loshchilov2018decoupled} as our optimizer and cross-entropy as the loss function. We used Google Colab\footnote{\url{https://colab.research.google.com/}} for our experiments.

\begin{table*}[t!]
\centering
\resizebox{\textwidth}{!}{%
\begin{tabular}{l|l|ccc|ccc|ccc|ccc|ccc}
\hline
 & \multicolumn{1}{c|}{} & \multicolumn{3}{c|}{\textbf{BLEU} (↑)} & \multicolumn{3}{c|}{\textbf{BLEURT} (↑)} & \multicolumn{3}{c|}{\textbf{CIDER} (↑)} & \multicolumn{3}{c|}{\textbf{CS} (↑)} & \multicolumn{3}{c}{\textbf{PPL} (↓)} \\ \cline{3-17} 
\multirow{-2}{*}{\textbf{Fine-tuned on}} & \multicolumn{1}{c|}{\multirow{-2}{*}{\textbf{Model}}} & \textbf{Test-S} & \textbf{Test-K} & \textbf{Chart-To-Text} & \textbf{Test-S} &\textbf{ Test-K} & \textbf{Chart-To-Text} & \textbf{Test-S} & \textbf{Test-K} &\textbf{ Chart-To-Text} &\textbf{Test-S} & \textbf{Test-K} &\textbf{ Chart-To-Text} &\textbf{Test-S} & \textbf{Test-K} & \textbf{Chart-To-Text}
\\ \hline
 & T5-Base & 18.06 & 8.89 & 14.98 & -0.2466 & -0.9940 & -0.0969 & 2.735 & 1.143 & 3.239 & 68.86 & 55.2 & 67.2 & 10.8742 & 8.2657 & 11.3778 \\
 & BART-Base & 13.3 & 6.79 & 12.99 & -0.0483 & -0.4208 & 0.0080 & 3.028 & 1.304 & 3.283 & 63.03 & 55.43 & 66.1 & 11.0813 & 9.4579 & 9.9658 \\
 & BART-Large-CNN & 22.79 & 6.34 & 22.98 & 0.0685 & -0.5161 & 0.0462 & 2.923 & 1.051 & 3.111 & 83.76 & 66.49 & 85.41 & 9.9658 & 6.7668 & 8.248 \\
\multirow{-4}{*}{Chart to Text \cite{kanthara2022chart}} & BART-Large-XSUM & 22.18 & 6.58 & 21.35 & 0.0496 & -0.4798 & 0.0259 & 2.866 & 1.067 & 2.973 & 83.23 & 65.61 & 85.15 & 9.2262 & 6.996 & 8.1973 \\ \hline
 & T5-Base & 26.86 & 48.84 & 19.15 & 0.1942 & 0.2400 & 0.0867 & 3.82 & 4.149 & 3.136 & 77 & 77.25 & 71.23 & \textbf{8.1932} & \textbf{4.9536} & 8.9733 \\
 & BART-Base & 11.69 & 21.23 & 11.58 & -0.0265 & -0.0259 & -0.0305 & 3.479 & 3.073 & 3.134 & 65.16 & 70.04 & 66.61 & 10.0825 & 10.5109 & 11.7486 \\
 & BART-Large-CNN & 29.91 & 24.37 & 22.87 & 0.1869 & -0.0276 & 0.0910 & 3.401 & 2.614 & 2.992 & 85.33 & 76.59 & \textbf{84.52} & 8.7723 & 11.6479 & 9.3914 \\
\multirow{-4}{*}{ChartSumm} & BART-Large-XSUM & 31.73 & 26.15 & 24.94 & 0.1963 & 0.0030 & 0.1107 & 3.551 & 2.718 & 3.092 & \textbf{85.58} & 77.2 & 84.02 & 9.1612 & 11.262 & 9.1076 \\ \hline
 & T5-Base & 10.1 & \textbf{50.74} & 7.6 & -0.2283 & 0.2624 & -0.2753 & 2.047 & 4.31 & 1.964 & 61.36 & 77.63 & 55.38 & 14.1075 & 5.1099 & 12.8642 \\
 & BART-Base & 7.29 & 37.57 & 6.79 & -0.1128 & 0.2255 & -0.1471 & 1.443 & 3.865 & 1.431 & 54.62 & 77.17 & 53.13 & 10.6306 & 9.0209 & 8.7061 \\
 & BART-Large-CNN & 10.31 & 29.83 & 10.21 & -0.2537 & 0.0085 & -0.2546 & 1.594 & 3.036 & 1.667 & 69.62 & \textbf{81.24} & 72.22 & 24.3287 & 12.2263 & 15.1434 \\
\multirow{-4}{*}{ChartSumm-k} & BART-Large-XSUM & 7.12 & 43.27 & 6.12 & -0.1569 & \textbf{0.3416} & -0.2128 & 1.672 & \textbf{4.422} & 1.638 & 58.72 & 80.06 & 56.45 & 18.7177 & 8.5596 & 14.015 \\ \hline
 & T5-Base & 26.84 & 14.39 & 21.11 & 0.1970 & -0.1449 & 0.1328 & 3.811 & 1.656 & 3.435 & 76.98 & 62.95 & 73.8 & 8.2032 & 5.6382 & \textbf{7.9724} \\
 & BART-Base & 21.93 & 10.23 & 17.25 & 0.1582 & -0.1622 & 0.0978 & 3.349 & 1.341 & 3.219 & 74.78 & 62.93 & 71.7 & 9.4447 & 6.2375 & 9.7854 \\
 & BART-Large-CNN & \textbf{32.27} & 7.6 & \textbf{28.71} & 0.1998 & -0.3784 & 0.1628 & 3.563 & 1.101 & 3.235 & 85.33 & 66.97 & 83.6 & 9.4005 & 6.9521 & 9.5459 \\
\multirow{-4}{*}{ChartSumm-s} & BART-Large-XSUM & 31.67 & 10.67 & 25.42 & \textbf{0.2832} & -0.1178 & \textbf{0.2377} & \textbf{3.929} & 1.271 & \textbf{3.563} & 81.26 & 65.2 & 78.97 & 9.315 & 5.9528 & 9.572 \\ \hline
\end{tabular}%
}

\caption{Experimental results on different test sets. Here, ‘↑’ means ‘Higher’ and ‘↓’ means ‘Lower’ value is better.}

\label{tab:evaluation-results}
\end{table*}

\begin{table*}[t!]
\centering
\resizebox{\textwidth}{!}{%
\begin{tabular}{c | p{0.3\textwidth} | p{0.3\textwidth} | p{0.2\textwidth}}
\hline
    \raisebox{-\totalheight}{\includegraphics[width=0.35\textwidth]{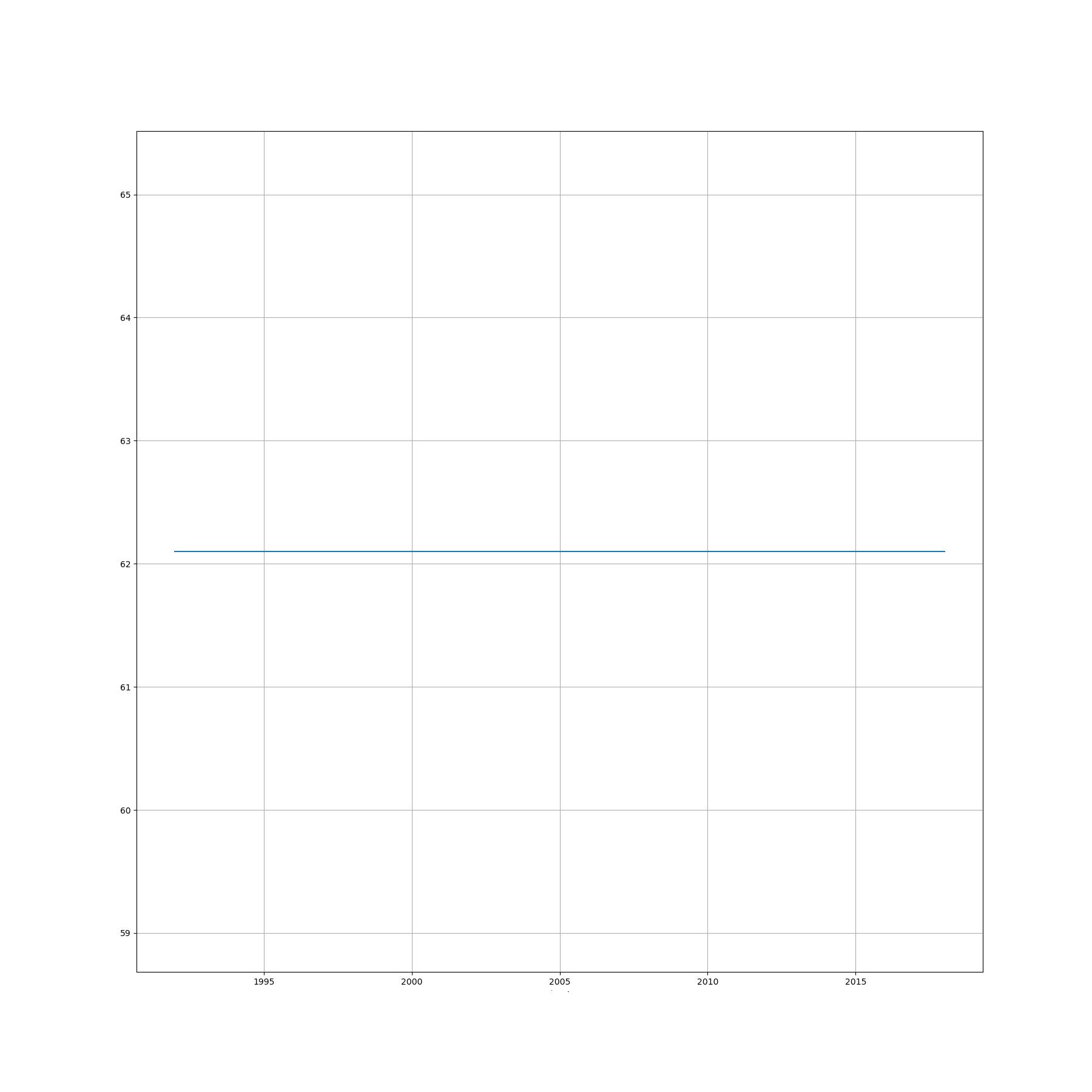}}
    & \textbf{Gold:} Between 1992 and 2018, Georgia renewable surface water remained stable at around 62.1 billion cubic meters per year.
    & \textbf{T5-Base:} In 2018 , renewable surface water for Georgia was \color{red}62.1 million cubic meters \color{black} per year. \color{orange} Though Georgia renewable surface water fluctuated substantially in recent years , it tended to increase through 1997 - 2018 period ending at 62.1 million cubic meters per year in 2018 . \color{black}
    & \textbf{Comments:} Model predicts factually incorrect information and captures incorrect trend. \\ \hline
    
    \raisebox{-\totalheight}{\includegraphics[width=0.35\textwidth]{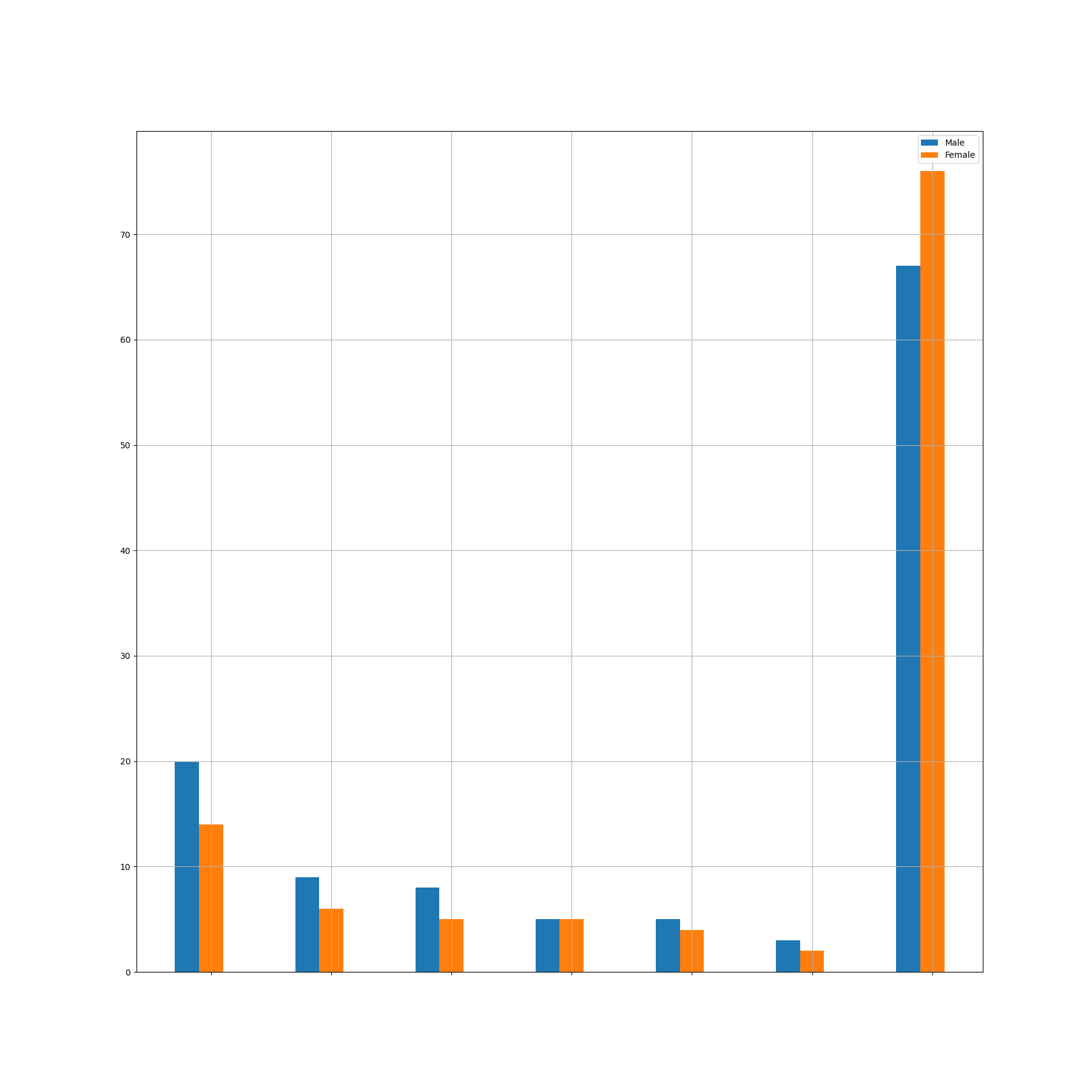}}
    & \textbf{Gold:} A survey from April 2020 revealed that during the coronavirus (COVID-19) lockdown about three out of ten Italian adults subscribed to paid digital services, with slightly higher penetration rates among male respondents. 20 percent of male adults subscribed to video on demand platforms, while just 14 percent of female respondents did the same.
    
    & \textbf{BART-Base:} This statistic shows the number of paid digital subscriptions during the coronavirus outbreak in Italy in 2020, by user gender.     
    & \textbf{Comments:} Model fails to generate informative summary. \\ \hline
    
    \raisebox{-\totalheight}{\includegraphics[width=0.35\textwidth]{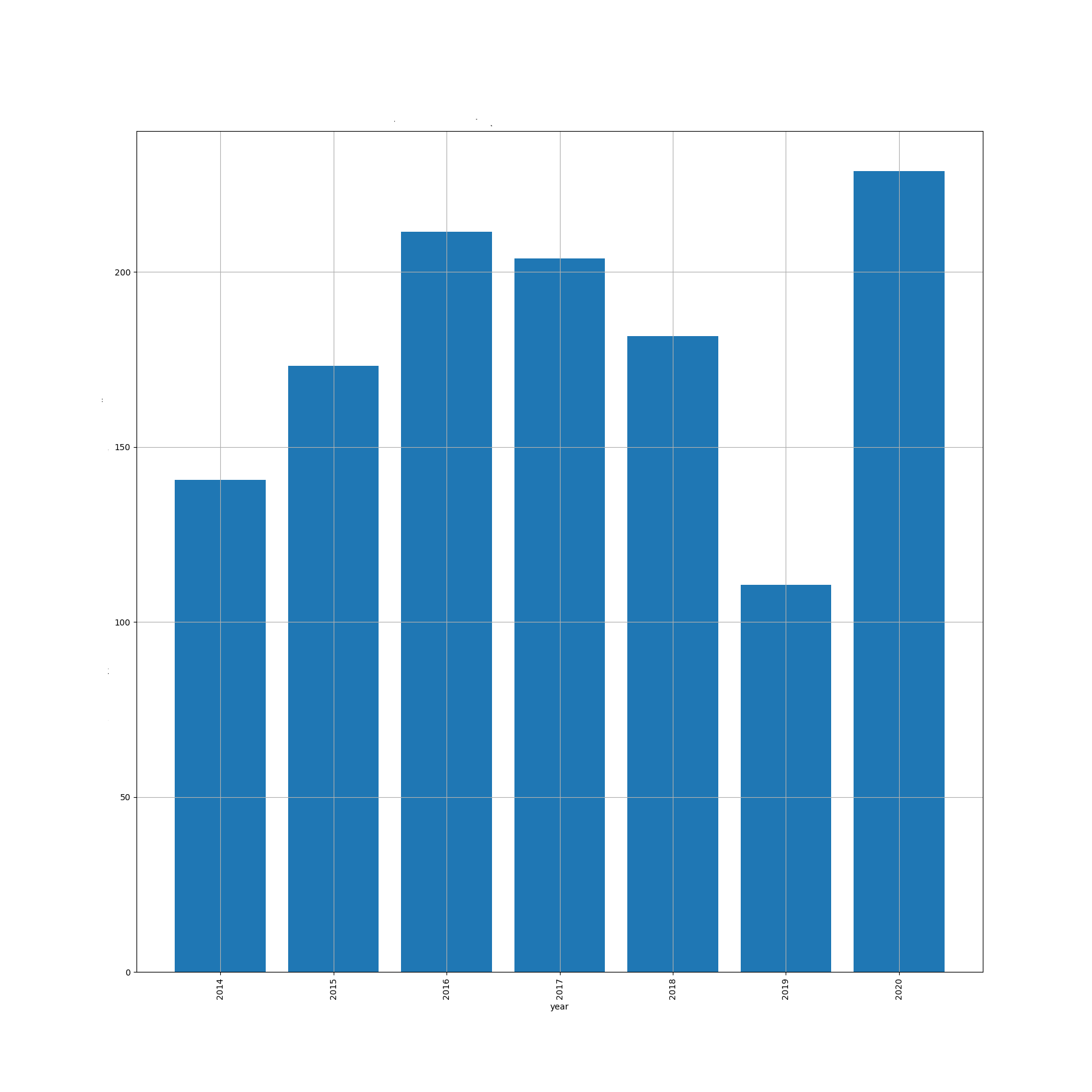}}
    & \textbf{Gold:} This statistic depicts the gross profit of Nautilus Inc. worldwide from 2014 to 2020 . In 2020 , Nautilus Inc. had a gross profit amounting to just under 230 million U.S. dollars , which is a significant increase compared to 2019 . &
    
    \textbf{BART-Large-CNN:} The statistic shows the gross profit of Nautilus Inc. worldwide from 2014 to 2020 . In 2020 , the company 's global gross profit amounted to approximately 228.8 million U.S. dollars . \color{violet} The company is headquartered in Seattle , Washington , United States . \color{black} &
    \textbf{Comments: } Model hallucinated about its headquarter. \\ \hline
\end{tabular}%
}
\caption{Examples of generated summaries containing error. \color{red} Red \color{black} indicates factually incorrect output, \color{orange} orange  \color{black} indicates failure to capture trend and \color{violet} violet \color{black} indicates hallucination error}
\label{generatedSummary}
\end{table*}

%\section{Baseline Architecture}
%\label{APPENDIX:architecture}

%\subsection{Hyperparameters}
%\label{APPENDIX:hyperparam}

%Hyperparameter and fine-tune process of our baseline is given in \ref{APPENDIX:architecture}
 %

\subsection{Evaluation Metrics:} We use five evaluation metrics in our automated evaluation: (i) BLEU \citep{post-2018-call}: it uses n-gram overlaps between reference text and machine-generated text to determine similarity score, (ii) BLEURT \citep{sellam-etal-2020-bleurt}: it evaluates how fluent the candidate is and how well it transfers the reference's meaning (we utilize BLEURT base-128 for our evaluation), (iii) Perplexity: it is a measurement that quantifies how well a probability model predicts a sample (we utilized pre-trained GPT-2 \citep{radford2019language} to measure perplexity), (iv) CIDEr: \citep{Vedantam_2015_CVPR} it uses n-grams overlaps and calculates average cosine similarity between the candidate sentence and the reference sentences, to capture the grammatical qualities with richer semantics, (v): Content Selection (CS): it measures how closely the generated text matches the reference documents \citep{wiseman2017challenges}.

\begin{table*}[t!]
\centering
\resizebox{0.75\textwidth}{!}{%
\begin{tabular}{l|ccc|ccc}
\hline
\multirow{2}{*}{\textbf{Fine-tuned On}} & \multicolumn{3}{c|}{\textbf{Test-k}} & \multicolumn{3}{c}{\textbf{Test-s}} \\ \cline{2-7} 
                               & \textbf{BLEU-1}  & \textbf{BLEU-2}  & \textbf{BLEU-N}  & \textbf{BLEU-1}  &\textbf{ BLEU-2}  & \textbf{BLEU-N} \\ \hline
Full dataset                   & 17.07   & 6.41    & 3.22    & 28.45   & 12.28   & 7.69   \\ \hline
Knoema                         & \textbf{17.80}  & \textbf{6.62}    & \textbf{3.79}    & 12.02   & 3.60    & 1.25   \\ \hline
Statista                       & 15.44   & 5.02    & 1.73    & \textbf{28.93}   & \textbf{12.63}   & \textbf{7.97}   \\ \hline
\end{tabular}%
}
\caption{Experimental results on different test sets of ChartSumm translated to Bengali.}
\label{evaluation-in-bengali}
\end{table*}

\subsection{Experimental Results:} To evaluate the performance, we use three test sets: (i) \textit{test-k}: denotes ChartSumm test set compiled from Knoema where the summaries are precise and well structured, (ii) \textit{test-s}: denotes ChartSumm test set compiled from Statista where the summaries are descriptive and have a longer length, and (iii) \textit{Chart-To-Text (Statista)}: the Statista version of the test set of Chart-To-Text \cite{kantharaj-etal-2022-chart}. We run experiments using the full training sets of both ChartSumm and Chart-To-Text, with additional experiments with the training subsets of Statista and Knoema from ChartSumm. 

Our experimental results are shown in Table \ref{tab:evaluation-results}. In terms of BLEURT, CIDER, and CS metrics, we observe that BART-Large models (Bart-Large-CNN and Bart-Large-XSUM) outperform other models in all three test sets, while T5 performs the best in terms of PPL in all test sets. Furthermore, we observe that models fine-tuned on ChartSumm-K always perform better than other models in test sets containing data from Knoema only (except the PPL metric). We again observe similar trends in terms of models fine-tuned on ChartSumm-S. More importantly, all models fine-tuned on our ChartSumm-S dataset outperform the baselines that are fine-tuned on Chart-To-Text even in the Chart-To-Text test set, showing the effectiveness of our proposed dataset. Meanwhile, models fine-tuned on Chart-To-Text performs very poorly in ChartSumm-K (about 82.48\% lower in terms of best performing T5 model), indicating that the Chart-To-Text dataset is not generalizable to generate summaries that are shorter and precise. To further investigate the performance, we do some error analyses in the following section. 

\subsection{Error Analysis and Challenges:}
For error analysis, we randomly sampled $100$ instances with their summaries generated by different baseline models.

We notice that in many cases, even though the generated summary is fluent and readable, it contains factually incorrect information and predicted wrong trend in data (see the first example in Table \ref{generatedSummary}. We also notice that models sometimes fail to generate informative summaries while also failing to predict anything about data (see the second example in Table \ref{generatedSummary}). In both Bart-Large models, we also find some cases where the models generate information that is fully irrelevant to the chart (i.e., the Hallucination effect \citep{gong2019enhanced, obeid-hoque-2020-chart, wiseman-etal-2021-data}) (see the third example in Table \ref{generatedSummary}).

\subsection{Evaluation of ChartSumm in Other Languages}

To assess the potential for expanding the use of ChartSumm to other languages, we undertook a study where we translated ChartSumm into Bengali and fine-tuned a pre-trained mT5 \cite{xue-etal-2021-mt5} model to evaluate its performance. To our best knowledge, this is the first study that has explored the task of summarizing charts in languages other than English.

\textbf{Translation:} To translate the training and validation sets of ChartSumm into Bengali, we utilized NLLB \cite{costa2022no}, which is a state-of-the-art neural machine translation model. To ensure proper evaluation, the test data was translated into Bengali with the assistance of human annotators who are undergraduate students and have proficiency in both English and Bengali. 

\textbf{Baseline:} In our study, we employed mT5 \cite{xue-etal-2021-mt5} as a baseline model due to its efficacy in text summarization tasks in Bengali. We fine-tuned a variant of mT5 which was pre-trained on multilingual XL-SUM \cite{hasan-etal-2021-xl}. Our baseline model was fine-tuned for 4 epochs using a batch size of 8, with an initial learning rate of 0.000001. The AdamW optimizer was used for the fine-tuning process and cross-entropy was utilized as the loss function. 

\textbf{Evaluation:} In our study, we conducted automatic evaluation of the text summarization models using the BLEU, which measures the degree of similarity between the reference and the machine-generated text using n-gram overlaps. However, we were unable to employ other evaluation metrics such as BLEURT and CIDEr for the Bengali language, as they require language-specific models that are not available for Bengali. Therefore, we only used BLEU as an evaluation metric for our Bengali text summarization experiment. We present the outcomes of our experiments in Table \ref{evaluation-in-bengali}. It is evident from the results that models fine-tuned on ChartSumm-s and ChartSumm-k performed better on their corresponding test sets in comparison to the scenario when evaluated on the combined test set. From this, we can see that the model can perform better in the respective test sets even with machine-generated translations. This opens up the possibility to investigate the performance of ChartSumm in other languages through automatic machine translations. Meanwhile, performance evaluation via human-annotated training data in other languages is also something worth investigating in the future.  

\section{Conclusion}

In this work, we present a new large scale benchmark dataset for the automatic chart summarization task to address the low resource problem in such tasks. Our proposed dataset is almost double in size than the existing largest dataset available for this task. Thus, the proposed ChartSumm dataset will serve as a strong benchmark for researchers in this relatively new area of natural language generation. We utilize three BART models and one T5 model as baselines and conduct extensive experiments using various evaluation metrics to identify the challenges in this task. Experimental results showed that models fine-tuned with our proposed ChartSumm dataset could achieve better domain generalization than other existing benchmark datasets. We also explored the possibility of extending ChartSumm to other languages through automatic machine translation. In the future, we would like to extend ChartSumm to a multilingual dataset to address the scarcity of well-formatted datasets in other low-resource languages. We will also study how to incorporate query relevance \cite{laskar-etal-2022-domain,laskar2020query,laskar-etal-2020-wsl}, question-answering \cite{masry2022chartqa,kantharaj2022opencqa,hoque2022chart,laskar2020contextualized}, and entity recognition \cite{laskar-etal-2022-blink,laskar2022auto,laskar2022improving} capabilities in this task.

\bibliography{references, custom}
\bibliographystyle{acl_natbib}

\end{document}